\documentclass[10pt,twocolumn,letterpaper]{article}
\usepackage[pagenumbers]{cvpr} 

\usepackage{adjustbox}
\usepackage{pifont}
\usepackage{amsmath} 

\newcommand{\cmark}{\ding{51}}%
\newcommand{\xmark}{\ding{55}}%

\newcommand{\dname}{WeatherProof Dataset}

\usepackage[dvipsnames]{xcolor}

\toggletrue{cvprfinal}

\definecolor{cvprblue}{rgb}{0.21,0.49,0.74}
\usepackage[pagebackref,breaklinks,colorlinks,citecolor=cvprblue]{hyperref}

\usepackage[pagenumbers]{cvpr} 

\usepackage{graphicx}
\usepackage{amsmath}
\usepackage{amssymb}
\usepackage{booktabs}
\usepackage{makecell}
\usepackage{multirow} 
\usepackage[ruled,longend]{algorithm2e} 
\usepackage{pythonhighlight} 
\usepackage{blindtext}
\usepackage{enumerate}
\usepackage{subcaption}
\usepackage{adjustbox}
\usepackage{pifont}
\usepackage[pagebackref,breaklinks,colorlinks]{hyperref}

\usepackage[pagebackref,breaklinks,colorlinks]{hyperref}

\usepackage[capitalize]{cleveref}
\crefname{section}{Sec.}{Secs.}
\Crefname{section}{Section}{Sections}
\Crefname{table}{Table}{Tables}
\crefname{table}{Tab.}{Tabs.}
\begin{document}

\title{WeatherProof: A Paired-Dataset Approach to Semantic Segmentation in Adverse Weather}




\author{Blake Gella$^1$\thanks{Equal contribution.} \quad Howard Zhang$^1$\footnotemark[1] \quad  Rishi Upadhyay$^1$ \quad Tiffany Chang$^1$ \quad Matthew Waliman$^1$ \\\quad Yunhao Ba$^1$\quad Alex Wong$^2$ \quad Achuta Kadambi$^1$\\
\normalsize{$^1$University of California, Los Angeles \quad
$^2$Yale University}}

\maketitle
\begin{abstract}
The introduction of large, foundational models to computer vision has led to drastically improved performance on the task of semantic segmentation. However, these existing methods exhibit a large performance drop when testing on images degraded by weather conditions such as rain, fog, or snow. We introduce a general paired-training method that can be applied to all current foundational model architectures that leads to improved performance on images in adverse weather conditions. To this end, we create the \dname, the first semantic segmentation dataset with accurate clear and adverse weather image pairs, which not only enables our new training paradigm, but also improves the evaluation of the performance gap between clear and degraded segmentation. We find that training on these paired clear and adverse weather frames which share an underlying scene results in improved performance on adverse weather data. With this knowledge, we propose a training pipeline which accentuates the advantages of paired-data training using consistency losses and language guidance, which leads to performance improvements by up to 18.4\% as compared to standard training procedures.


\end{abstract}
\vspace{-10pt}    
\section{Introduction}
\label{sec:intro}
Semantic segmentation has a rich history due to its countless applications in autonomous driving~\cite{ess2009segmentation,nekrasov2019real,siam2018rtseg,zhao2018icnet}, robotics~\cite{kim2018indoor,milioto2018real,milioto2019bonnet}, and scene understanding~\cite{gupta2015indoor,cordts2016cityscapes,li2009towards}. The current pace of advancements has been accelerated by the introduction of large, generally-pretrained, foundational models. In fact, these foundational models consistently dominate the leaderboards of competitive semantic segmentation benchmarks, i.e. ADE20K~\cite{zhou2017scene}, Cityscapes~\cite{cordts2016cityscapes}, by placing in the top 5 rankings. Yet, despite their success on these leaderboards, when presented with images with visual degradations, i.e., those captured under adverse conditions, their performance similarly degrades.

\begin{figure}
    \centering
    \includegraphics [width=\linewidth]{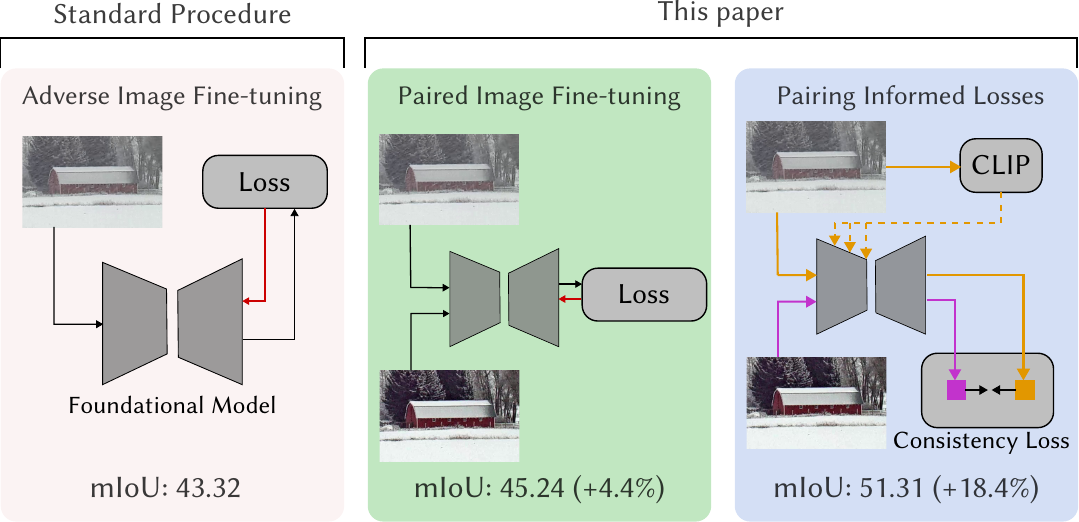}
    \caption{\textbf{By leveraging the full paired-training method, we improve InternImage's performance on adverse weather conditions by up to 18.4\%.}}
    \label{fig:teaser}
\end{figure}

The real-world effects of adverse weather conditions such as rain, fog, or snow have been shown by many previous studies to have very complex visual degradation patterns\cite{wang2019spatial,ba2022not,zhang2023weatherstream,liu2018desnownet,tan2008visibility} -- patterns that can be affected by factors including but not limited to atmospheric condition, camera parameters, or even geographic location\cite{zhang2023weatherstream,ba2022not}. The prevalence of weather in our natural world translates to performance gaps that arise in our algorithms. Existing work~\cite{sakaridis2018semantic,sakaridis2021acdc,kerim2022semantic} has provided datasets and methods with the goal of studying the effects of these natural phenomenons. However, as it is difficult to capture paired datasets to study them in a controlled setting, existing datasets have resorted to the use of synthetic weather effects, or include mis-alignments in the underlying scene between adverse and clear-weather images. To address this, we build off of the WeatherStream dataset~\cite{zhang2023weatherstream} to introduce the \dname, the first semantic segmentation dataset with accurately paired clear and weather-degraded image pairs. By ensuring the underlying semantic labels are the same between clear and adverse weather images, we provide a controlled test bench where performance degradations can be largely isolated to weather artifacts.

Using this dataset, we tested several modern foundational model architectures, which underperformed on adverse-weather images compared to clear-weather images (see~\cref{table:Degraded-table}). Motivated by this study, we propose a paired-training method to address the limitations of foundational models under adverse conditions. The crux of our method is that when we fine-tune a generally pre-trained foundational model for semantic segmentation in adverse weather, we observed two trends which the model seeks to reduce the loss. It either (1) learns to adapt to new scenes, or (2) learns to adapt to weather degradations within the same scene. By training on images with adverse weather only, or by training on both clear and adverse images that do not share an underlying scene structure, we entangle (1) and (2), forcing the model to learn both at the same time. In general, inter-scene variance can arise from changes in a number of factors, including camera parameters, resolution, environment, etc. Thus, inter-scene variance is usually much higher than changes within the same scene due to weather degradations. Therefore, the model generally focuses on learning to adapt to new scenes (1). This is empirically validated in~\cref{sec:exp_grad}. Because the model learning prioritizes (1), learning both simultaneously may lead to a performance drop on adverse weather images. We find in this work that by training on a paired dataset such as \dname, where the underlying scene and segmentation labels are the same, we ease the training process by allowing the model to focus more on (1) when training on clear images, and focus more on on (2) when training on adverse images (since the model has trained on the same scene without weather degradations already).

While this training scheme already results in improvements in adverse weather performance, we find that the model still prioritizes (1) over (2). To reduce this priority gap further, we leverage consistency losses and language guidance. We propose a Feature Consistency Loss (FCL) and an Output Consistency Loss (OCL). These losses take advantage of the fact that in the \dname, the underlying scene shares the same semantic features between a clear and adverse weather image. By pulling both the extracted features (FCL) as well as the outputs (OCL) together for clear and adverse weather image inputs, we push the model architecture to learn a latent space less susceptible to weather degradations, which, unlike just cross entropy loss, prioritizes learning to adapt to weather degradations. 

We can additionally focus the model on learning to adapt to weather degradations by injecting information about the weather condition into the model. The overall weather in a scene is usually a composition of multiple weather effects such as rain, fog, or snow. Knowledge of that composition could be beneficial to the model for becoming resilient to weather by limiting the search space of possible feature representations that it has to consider. To do this, we propose a CLIP Injection Layer, which uses natural language to guide the model by estimating this composition and injecting it into the network through cross attention.

By allowing the model to separately learn adapting to new scenes and adapting to the addition of weather degradations through the use of both consistency losses as well as the CLIP Injection Layer, our paired-training method obtains relative increases of up to 18.4\% mIOU as compared to standard training procedures on our new \dname\ evaluation set.

\subsection{Contributions}

In summary, we make the following contributions:
\begin{itemize}[itemsep=1pt]
    \item We introduce the \dname\, a semantic segmentation dataset with over 174.0K images. It is the first with high quality semantic segmentation labels with accurately paired clear and adverse weather images for paired training and more accurate evaluation. Training on this paired dataset decouples the task of learning new scenes and learning resiliency to weather effects, improving model performance on weather-degraded scenes.
    
    \item We augment the paired training process by utilizing a Feature Consistency Loss and Output Consistency Loss, which focuses the model on learning resiliency to weather effects, improving performance on adverse weather conditions.
    
    \item We further augment the paired training process through a CLIP Injection Layer, which helps the model by injecting the composition of the weather effect through cross attention and language guidance, improving performance on adverse weather conditions.

    \item Together, the full paired data training process leads to an improvement of up to 18.4\% on adverse weather-degraded images.
    
\end{itemize}
\section{Related Works}
\label{sec:related}
\subsection{Vision Foundation Models}
Recently, research has shown the superb performance of deep learning models such as CNNs~\cite{hong2015decoupled,hu2018learning,lin2017refinenet,long2015fully} or vision transformers~\cite{xie2021segformer,hu2021istr}. With the rising popularity of these convolution and attention-based architectures, a recent wave of research has shown the excellent learning capability and segmentation performance of many different foundational models. The Swin architecture uses a shifted window technique to achieve both local and global attention while maintaining linear computational complexity with respect to image size~\cite{liu2021swin, liu2022swin}. The ConvNeXt model makes multiple changes to the standard convolutional network training pipeline (kernel sizes, activations, etc.) to modernize the CNN approach to outperform the Swin~\cite{liu2022convnet}. The InternImage model uses deformable convolutions to maintain the long-range dependence of attention layers in a low memory/computation regime~\cite{wang2023internimage}. However, all of these aforementioned models benchmark their segmentation performance on the ADE20K dataset~\cite{zhou2017scene}, which does not contain images in adverse conditions, such as adverse weather. As such, the robustness of these foundational models are yet to be evaluated.

\begin{table}[t]
    \begin{adjustbox}{width=\columnwidth}
    \centering
    \begin{tabular}{ccccc}
    \toprule
    Dataset & $\#$ Images & Weather Degradations? & Real? & Paired? \\
    \midrule
    ADE20K~\cite{zhou2017scene} & 27.5K & \xmark & \cmark & \xmark \\
    Cityscapes~\cite{cordts2016cityscapes} & 25K & \xmark & \cmark & \xmark \\
    Foggy Cityscapes~\cite{sakaridis2018semantic} & 35K & \cmark & \xmark & \xmark \\
    ACDC~\cite{sakaridis2021acdc} & 4K & \cmark & \cmark & \xmark \\
    \midrule
    \dname\ (Ours) & 174.0K & \cmark & \cmark & \cmark \\
    \bottomrule
    \end{tabular}
    \end{adjustbox}
\caption{\textbf{\dname\ is the first high quality annotated segmentation dataset with accurate clear and weather-degraded image pairs for better consistency loss in training and evaluation.} Other datasets either do not contain adverse weather effects, have synthetic weather effects, or do not have accurate paired clear images. While ACDC does have paired images, there exists a stark difference between the adverse and reference images, which is shown in~\cref{fig:dataset}} 
\label{tab:related_work}
\end{table}

\subsection{Semantic Segmentation in Adverse Weather}
\label{sec:related_adverse}
The two most popular semantic segmentation datasets are ADE20K~\cite{zhou2017scene} and Cityscapes~\cite{cordts2016cityscapes}. ADE20K does not include images with adverse conditions, and Cityscapes avoids the adverse weather condition as well. Efforts have since been made to provide adverse conditions through synthetic means, such as the generation of the Foggy Cityscapes dataset~\cite{sakaridis2018semantic}. However, research in the past has shown that there exists a performance gap when training on synthetic weather conditions~\cite{ba2022not,zhang2023weatherstream}. The ACDC dataset provides real images with segmentation labels with weather conditions present~\cite{sakaridis2021acdc}. They also provide a paired frame for each adverse frame representing the clear weather version. However, it can be seen in~\cref{fig:dataset} that there are inconsistencies in the pairs that rule out its effectiveness in paired training approaches. A comparison of segmentation datasets can be seen in~\cref{tab:related_work}.

\subsection{Language and Vision}
\label{sec:related_language}
The domains of language and vision have often been separate in machine learning. Recent works have begun to incorporate language into vision models~\cite{radford2021learning,li2022blip,rombach2022high,nichol2021glide}. The BLIP model trains a captioner and filter and uses it on Internet images to achieve high performance results on vision-language tasks~\cite{radford2021learning}. The CLIP model learns a shared latent space between image and text by contrastive pretraining on image-text pairs. CLIP has shown to be fundamental for enabling vision models with language priors, as seen with GLIDE~\cite{nichol2021glide} and Stable Diffusion~\cite{rombach2022high}. Stable Diffusion uses CLIP's text encoder to inject prompts into their UNET's cross attention layers, guiding image generation through text. However, vision models have yet to utilize language for guidance through adverse conditions such as weather.
\section{Methods}
\begin{figure*}[t]
    \centering
    \includegraphics [width=\linewidth]{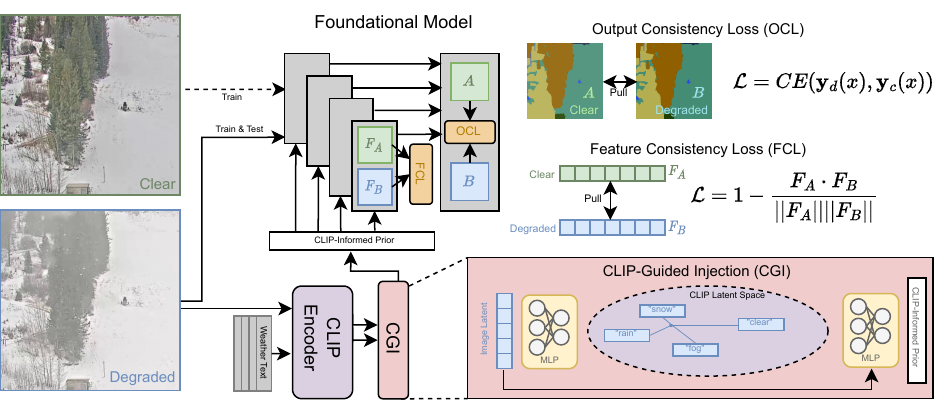}
    \caption{\textbf{By using a paired-training method with consistency losses and CLIP injection, foundational models are able to generate features that are more resilient to adverse weather conditions.} During paired data training, a CLIP-Guided Injection module learns a CLIP-informed prior representing the adverse weather effect in the CLIP latent space. Clear and adverse weather images are fed into a shared weight encoder-decoder structure. Intermediate features and output segmentation maps are used in a feature consistency loss and an output consistency loss respectively to ensure an advantageous representation.}
    \label{fig:model}
\end{figure*}

In order to improve the performance of semantic segmentation models in the presence of adverse weather, we introduce the paired-data training method made possible by our newly introduced~\dname. In~\cref{sec:methods_image}, we go over the forward image model as used in past works regarding adverse weather conditions. In~\cref{sec:methods_dataset}, we describe the dataset used for paired image training. In~\cref{sec:methods_loss}, we define the consistency losses used to guide the model towards feature representations that perform well in adverse weather. In~\cref{sec:methods_clip}, we formalize the method used to inject CLIP-based language guidance into our network. See~\cref{fig:model} for an overview of all of the above.

\subsection{Image Formation Model}
\label{sec:methods_image}
In order to study and alleviate the performance gap faced by semantic segmentation models in weather conditions such as rain, fog, or snow, it is important for us to mathematically formalize how an image can be affected by different weather phenomenon. To do this, we initially borrow the light transport model proposed by research in the field of weather removal~\cite{ba2022not,zhang2023weatherstream,deng2018directional,fu2017removing,li2018non,li2019heavy,li2020all,li2016rain,valanarasu2022transweather,wang2020model,wang2019spatial,yasarla2019uncertainty,zhang2018density,zhu2017joint,chen2020jstasr,chen2021all,liu2018desnownet,he2010single,tan2008visibility}. Weather in an image can largely be attributed to one of two effects, particle effects (raindrops or snowflakes) or scattering effects (rain accumulation, snow veiling, haze, fog, etc.). Weather particles can be modeled as a convex combination of the underlying clear scene and a map of the particles. This can be done for rain as follows:
\begin{equation} \label{eq:rain_model}
    \mathcal{D}_{\text{rain}}(\mathbf{J}(x)) = \mathbf{J}(x)(1-\mathbf{M}_r(x)) + \mathbf{R}(x)\mathbf{M}_r(x),
\end{equation}
where $x$ represents the spatial location within an image, $\mathcal{D}_{\text{rain}}$ represents a function that maps a clear image to one with rain particle effects, $\mathbf{J}(x)$ represents the clear image with no weather effects, $\mathbf{M}_r(x)$ represents a mask of the locations of rain particles, and $\mathbf{R}(x)$ represents a map of the rain streaks~\cite{ba2022not,zhang2023weatherstream,deng2018directional,fu2017removing,li2018non,li2019heavy,li2020all,li2016rain,valanarasu2022transweather,wang2020model,wang2019spatial,yasarla2019uncertainty,zhang2018density,zhu2017joint}. For snow, it is:
\begin{equation} \label{eq:snow_model}
    \mathcal{D}_{\text{snow}}(\mathbf{J(x)}) = \mathbf{J}(x)(1-\mathbf{M}_s(x)) + \mathbf{S}(x)\mathbf{M}_s(x),
\end{equation}
where $\mathcal{D}_{\text{snow}}$ and $\mathbf{M}_s$ represent their corresponding snow equivalents, and $\mathbf{S}(x)$ represents a chromatic aberration map of the snow particles~\cite{chen2020jstasr,chen2021all,liu2018desnownet}.

Scattering effects are modeled through the use of the scene radiance equation, which, evaluated at each pixel location, is
\begin{equation} \label{eq:radiance}
\begin{split}
    \mathcal{D}_{\text{fog}}(\mathbf{J(x))} & = \mathbf{J}(x)e^{-\int_0^{d(x)}\beta dl} + \int_0^{d(x)}L_\infty \beta e^{-\beta l}dl, \\
    & = \mathbf{J}(x)e^{-\beta d(x)} + L_\infty(1-e^{-\beta d(x)}),
\end{split}
\end{equation}
where $\mathcal{D}_{\text{fog}}$ represents a function mapping a clear image to one with scattering effects, $d(x)$ represents the distance from the observer at a pixel location $x$, $\mathbf{J}(x)$ represents the radiance of the underlying scene (the clear image), $\beta$ is an atmospheric attenuation coefficient (assumed to be constant throughout the scene), and $L_\infty$ is the radiance of the airlight~\cite{tan2008visibility,he2010single}. 

Images degraded by adverse weather can be affected by any combination of~\cref{eq:rain_model,eq:snow_model,eq:radiance}. Having an estimate of how much an image is affected by each of these particular weather phenomenons can be utilized by the model to limit the search space of possibly advantageous feature representations. This estimation process will be explained in~\cref{sec:methods_clip}.

\subsection{Dataset and Paired Data Training}
\label{sec:methods_dataset}
\begin{figure*}
    \centering
    \includegraphics [width=\linewidth]{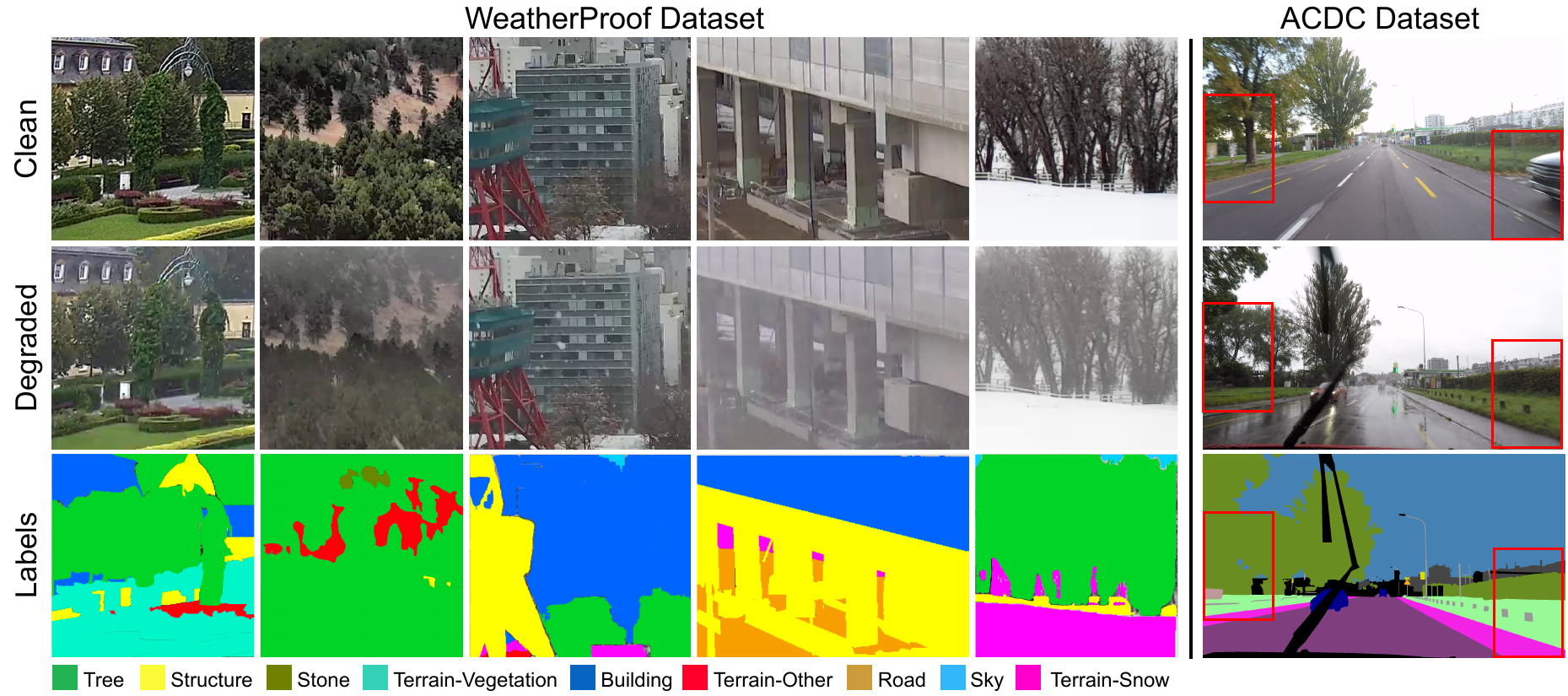}
    \caption{\textbf{\dname\ contains accurate clear and adverse weather image pairs with 10 semantic classes}. In contrast, the ACDC dataset's paired images have major differences in semantic information and scene structure.}
    \label{fig:dataset}
\end{figure*}
To train and evaluate our models, we use our own \dname\ containing 147.8K paired adverse and clear images for training, and 26.2K for testing. The clear and adverse image pairs were selected from the GT-RAIN~\cite{ba2022not} and WeatherStream~\cite{zhang2023weatherstream} datasets. This dataset was chosen due to its meticulous consideration of scene consistency between the clear and degraded images. By leveraging this scene consistency, we are able to construct the first semantic segmentation dataset with truly paired data. By paired data, we specifically mean data in which there are very minimal differences in the underlying clear scene as compared to the adverse one. That way, the two should have the exact same segmentation labels. In addition, we are also able to take advantage of the unique diversity of the WeatherStream dataset, containing geographic locations around the world in a variety of different urban and natural locations, with varying camera parameters and resolutions for dataset variety. We label the 10 following classes in the datasets: background, tree, structure, road, terrain-snow, terrain-grass, terrain-other, stone, building, and sky. To emphasize accurate object borders, we give priority to minimizing the amount of background labels between objects without mislabelling. Samples from the dataset can be seen in~\cref{fig:dataset}.

By creating an accurately paired semantic segmentation dataset, we are able to take advantage of a paired-data training pipeline. This simply involves feeding in both adverse images $\mathbf{I}_d(x)$ as well as clear images $\mathbf{J}(x)$ during training. By doing this, we allow the model to learn to adapt to new scenes when training on $\mathbf{J}(x)$, and adapt to weather degradations on the same scene when training on $\mathbf{I}_d(x)$.

On the testing side, the \dname\ allows one to get a more accurate evaluation of the exact performance gap that is present due to the presence of the degradations. Since the only differences (apart from extremely minute discrepancies) between two paired scenes is the presence of weather artifacts, any performance gap between the two is attributed to the adverse weather.

\subsection{Consistency Losses}
\label{sec:methods_loss}
As mentioned before, as a result of the inter-scene variance, learning the semantic labels of new scenes is much more conducive to minimizing the cross entropy loss than learning resilience to weather. Therefore, the model is more inclined to focus on the the former. To further alleviate this issue beyond simple paired data training, we guide the model towards a feature representation that is less susceptible to the adverse weather, which gives the model an extra incentive for learning resilience to weather, rather than solely focusing on minimizing the cross entropy loss with respect to the ground truth semantic labels. We take inspiration from the rain-invariant loss~\cite{ba2022not}, which aims to learn a latent space resilient against weather by utilizing a contrastive ``push/pull" between the clear and adverse latent features. Similarly, we propose two losses to improve the model's resilience to weather phenomenon: the feature consistency loss and the output consistency loss. Both losses take advantage of the \dname\ and its paired clear and adverse images. 

To formalize these losses, we propose two functions: $\mathcal{F}(\cdot,\theta_1)$, parameterized by $\theta_1$, that extracts features from an image and $\mathcal{G}(\cdot,\theta_2)$, parameterized by $\theta_2$, that produces a semantic segmentation map. Both functions are realized as neural networks. From here, we define 
\begin{align}
\hat{\mathbf{y}_d}(x) = \mathcal{G}(\mathcal{F}(\mathbf{I}_d(x),\theta_1),\theta_2), \\
\hat{\mathbf{y}_c}(x) = \mathcal{G}(\mathcal{F}(\mathbf{J}(x),\theta_1),\theta_2),
\end{align}
where $\hat{\mathbf{y}_d}(x)$ is the outputted semantic segmentation map in $\mathbf{\mathbb{R}}^{C \times H \times W}$ (channel, height, width) with the adverse weather image $\mathbf{I}_d(x)$ as the input, and $\hat{\mathbf{y}_c}(x)$ is the outputted semantic segmentation map with the clear image $\mathbf{J}(x)$ as the input. To train our model to become less susceptible to weather effects while extracting features, we implement the feature consistency loss that minimizes 
\begin{equation}
    \mathcal{L}_{\text{FCL}} = 1 - \dfrac{\mathcal{F}(\mathbf{J}(x),\theta_1) \cdot \mathcal{F}(\mathbf{I}_d(x),\theta_1)}{\|\mathcal{F}(\mathbf{J}(x),\theta_1)\| \|\mathcal{F}(\mathbf{I}_d(x),\theta_1)\|},
\end{equation}
where $\mathcal{L}_{feat}$ is the feature consistency loss. This pulls the extracted features of the clean image $\textbf{J}(x)$ and adverse image $\textbf{I}_d(x)$ together in the latent space. The feature consistency loss, thus, constrains the model feature extractor to find a representation that is less affected by weather effects.

We additionally propose an output consistency loss to minimize the following: 
\begin{align}
    \mathcal{L}_d &= CE(\hat{\mathbf{y}}_d(x), \hat{\mathbf{y}}_c(x)), \\
    \mathcal{L}_c &= CE(\hat{\mathbf{y}}_c(x), \hat{\mathbf{y}}_d(x)), \\
    \mathcal{L}_{\text{OCL}} &= \mathcal{L}_d + \mathcal{L}_c,
\end{align}
where $\mathcal{L}_d$ is cross entropy loss between the segmentation labels of the adverse weather images $\hat{\mathbf{y}}_d(x)$ and the detached segmentation labels of the clear images, $\mathcal{L}_c$ is the cross entropy loss between the segmentation labels of the clear images $\hat{\mathbf{y}}_c(x)$ and the detached segmentation labels of the adverse images, and $\mathcal{L}_{\text{OCL}}$ is the sum of both. The loss is calculated as above to maintain the validity of the backpropagation graph, while pulling the decoder features of both clear and adverse images together. Similar to the feature consistency loss, the output consistency loss encourages the model to remain resilient against weather when mapping from extracted features to the semantic segmentation map. Because adapting to new scenes accomplishes very little in minimizing either the feature consistency or output consistency, in order to learn this representation, the model must place additional focus on learning to adapt to weather degradations.

\subsection{CLIP Injection Layer}
\label{sec:methods_clip}
As alluded to earlier, estimating the contribution of weather phenomenons in an adverse image can help ease the model's ability to learn adaptability to different weather conditions. In order to accomplish this, we leverage language guidance. We utilize the CLIP model, which learns a latent space that is shared by both image and text encodings~\cite{radford2021learning}. To formalize the problem, we are given an adverse image $\mathbf{I}_d(x)$, which represents a clear image $\mathbf{J}(x)$ which has been affected by some combination of functions $\mathcal{D}_{\text{rain}},\mathcal{D}_{\text{snow}},\mathcal{D}_{\text{fog}}$, as well as a set of texts $\mathcal{T} = \{t_n\}_{n=1}^N$ describing $N$ different weather conditions. Our aim in this section is to find a vector ${\vec{v}} \in \mathbb{R}^N$ where each element represents the contribution of each weather effect to an adverse image.

\begin{figure}[t]
    \centering
    \includegraphics [width=\linewidth]{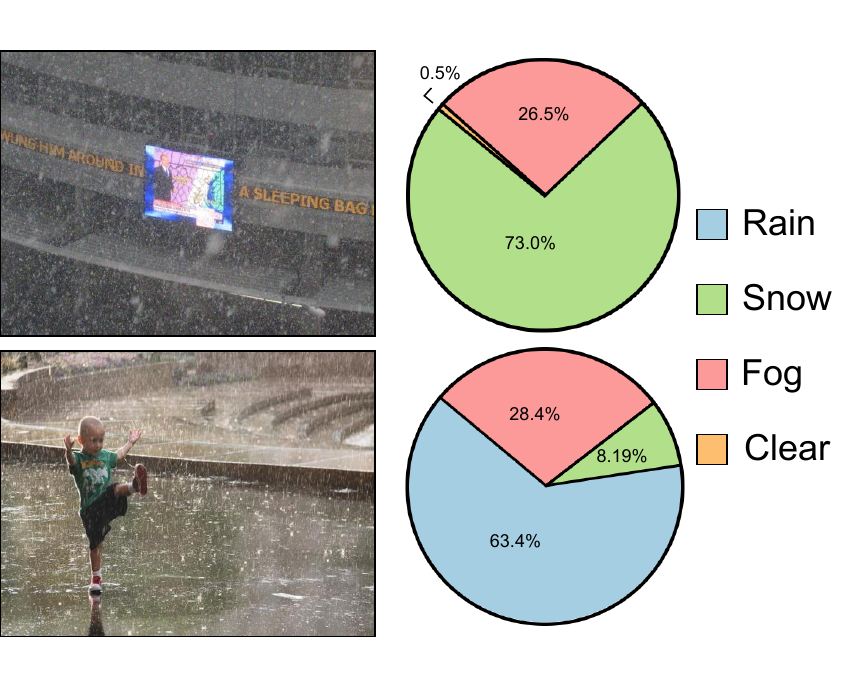}
    \caption{\textbf{Our CLIP injection layer is able to accurately predict the composition of weather effects in images.} The percentage of weather effect contributions was taken by passing in these images into our CLIP injection layer and extracting the weights $\vec{v}$.}
    \label{fig:clip_ref}
\end{figure}

We begin by passing $\mathbf{I}_d(x)$ as well as each text $t_n$ through a frozen clip encoder model to obtain CLIP embeddings in the shared latent space:
\begin{align}
    \vec{\mathbf{I}}_{\text{CLIP}} &= \text{CLIP}(\mathbf{I}_d(x)), \\
    \mathbf{T}_{\text{CLIP}} &= \{\text{CLIP}(t_n)\}_{n=1}^N,
\end{align}
where the CLIP$()$ function represents passing an image or text through the CLIP encoder, $\vec{\mathbf{I}}_{\text{CLIP}}$ represents the length $512$ feature vector representing the adverse weather image $\mathbf{I}_d(x)$, and $\mathbf{T}_{\text{CLIP}} \in \mathbb{R}^{N \times 512}$ represents the matrix of CLIP feature vectors representing the set of weather texts $\mathcal{T}$. We pass $\vec{\mathbf{I}}_{\text{CLIP}}$ through an MLP $f_\theta$ with parameters $\theta$ to obtain the $N$ length vector $\vec{v} \in \mathbb{R}^N$:
\begin{equation}
    \vec{v} = f_\theta(\vec{\mathbf{I}}_{\text{CLIP}}).
\end{equation}
To learn the parameters $\theta$ such that an accurate weight vector $\vec{v}$ can be learned, we first accumulate the text embeddings in the set $\mathbf{T}_{\text{CLIP}}$, weighted by $\vec{v}$, to obtain a final clip embedding $\vec{W} \in \mathbb{R}^{512}$ representing the unique weather condition present in the adverse scene $\mathbf{I}_d(x)$. This is concatenated with the clip embedding of the image $\mathbf{I}_\text{CLIP}$ to arrive at $\tilde{\mathbf{W}} \in \mathbb{R}^{1024}$:
\begin{align}
    \tilde{\mathbf{W}} &= \vec{W} \oplus\ \mathbf{I}_\text{CLIP}, \\
    \vec{W} &= \mathbf{T}_\text{CLIP} \cdot \vec{v}, 
\end{align}
where $\tilde{\mathbf{W}}$ is the final vector passed out of the layer into the rest of the model. To see examples of the learned weight vector $\vec{v}$ for unique scenes with different weather conditions, see~\cref{fig:clip_ref}.

\section{Experiments}
\label{sec:experiments}
We compare state-of-the-art foundational models to their counterparts trained with the paired-training method described above. All quantitative results use the IoU metric and are evaluated on our \dname. The models we train use pretrained checkpoints for their respective encoder. We use 13 text embeddings for the CLIP injection method. We implement a sigmoid loss schedule for our feature consistency loss and a step loss schedule for our output consistency loss. More details can be found in ~\cref{sec:Ablation}.

\subsection{InternImage}
We modified InternImage's~\cite{wang2023internimage} XL backbone by adding a cross-attention layer for CLIP Injections at the end of each stage. InternImage has a total of 39 layers in 4 stages, with stages ending at layers 5, 10, 34, and 39. Downsampling occurs after layers 5, 10, and 34. Before each downsampling layer of InternImage's encoder, a latent vector is outputted after a series of deformable convolution blocks, which is projected as queries and the CLIP injection $\tilde{\mathbf{W}}$ is projected as the keys and values to the cross-attention layer. We add a total of 3 cross-attention layers to InternImage.

\subsection{ConvNeXt}
ConvNeXt~\cite{liu2022convnet} has a total of 36 layers, with stages ending at layers 3, 6, 33, and 36. Downsampling occurs before layers 1, 3, 6, and 33. We add a cross-attention layer to ConvNeXt's XL backbone at the end of each stage. In total, 3 cross-attention layers are added.

\subsection{Results}
\label{sec:results}
\textbf{Results on Adverse Weather Images}: We see in~\cref{table:Degraded-table} that our paired training for both foundational model architectures results in improved performance by up to 4.4\% compared to the standard training procedure of fine-tuning on adverse images only. This is due to the model's increased ability to adapt to visual degradations from weather. We further improve this performance jump to 18.4\% by adding the CLIP Injection Layer and consistency losses.

\noindent
\textbf{Results on Clear Images}: Our focus on improving the performance on foundational architectures is due to its great performance in recent research on normal clear-weather data. We validate in~\cref{table:Clean-table} that our paired training method does not harm (and in fact benefits in some cases) the performance of these models on clear-weather data.

\noindent
\textbf{Adverse Weather Models}: We limit our scope in this paper to modern foundational models such as InternImage~\cite{wang2023internimage} and ConvNeXt~\cite{liu2022convnet} as they outperform previous methods. We validate this performance gap in~\cref{table:foundation}, where we test a model that is specifically designed to perform well on adverse weather conditions. We see that both ConvNeXt and InternImage drastically outperform by over 41\%.

\begin{table*}
  \centering
  \resizebox{\textwidth}{!}{
      \begin{tabular}{ccccccccccc}
        \toprule
        Model & \rotatebox{90}{Tree} & \rotatebox{90}{Struc.} & \rotatebox{90}{Road} & \rotatebox{90}{T-Snow} & \rotatebox{90}{T-Veg.} & \rotatebox{90}{T-Other} & \rotatebox{90}{Stone} & \rotatebox{90}{Building} & \rotatebox{90}{Sky} & mIoU $\uparrow$ \\
        \midrule
        InternImage Adverse Only~\cite{wang2023internimage} & 71.78 & 41.01 & \textbf{10.20} & 64.95 & 60.40 & \textbf{22.96} & 17.28 & 64.84 & 36.45 & 43.32 \\
        InternImage Paired~\cite{wang2023internimage}& 71.23 & 35.26 & 6.73 & \textbf{66.72} & 59.97 & 16.61 & 34.0 & 67.87 & 48.74 & 45.24\\
        InternImage Paired~\cite{wang2023internimage} + Losses + CLIP (Ours) & \textbf{74.73} & \textbf{46.71} & 6.60 & 66.55 & \textbf{64.8} & 19.33 & \textbf{50.07} & \textbf{73.99} & \textbf{58.98} & \textbf{51.31}\\
        \midrule
        ConvNeXt Adverse Only~\cite{liu2022convnet} & 66.4 & 45.83 & 7.66 & 45.43 & \textbf{58.67} & 14.62 & 24.69 & 59.45 & 37.88 & 40.07\\
        ConvNeXt Paired~\cite{liu2022convnet} & 62.32 & \textbf{53.34} & 5.20 & 51.14 & 53.70 & \textbf{16.33} & 20.69 & 60.69 & \textbf{44.69} & 40.92\\
        ConvNeXt Paired~\cite{liu2022convnet} + Losses + CLIP (Ours) & \textbf{68.74} & 39.63 & \textbf{7.80} & \textbf{56.10} & 57.77 & 15.21 & \textbf{40.41} & \textbf{69.33} & 40.26 & \textbf{43.92}\\
        \bottomrule
      \end{tabular}
    }
  \caption{\textbf{Our proposed paired training method outperforms standard fine-tuning on adverse images only for both InternImage~\cite{wang2023internimage} and ConvNeXt~\cite{liu2022convnet}.} Including language guidance and consistency losses further improve our results.}
  \label{table:Degraded-table}
\end{table*}

\begin{table*}
  \centering
  \resizebox{\textwidth}{!}{
      \begin{tabular}{ccccccccccc}
        \toprule
        Model & \rotatebox{90}{Tree} & \rotatebox{90}{Struc.} & \rotatebox{90}{Road} & \rotatebox{90}{T-Snow} & \rotatebox{90}{T-Veg.} & \rotatebox{90}{T-Other} & \rotatebox{90}{Stone} & \rotatebox{90}{Building} & \rotatebox{90}{Sky} & mIoU $\uparrow$\\
        \midrule
        InternImage Adverse Only~\cite{wang2023internimage} & 77.32 & 30.18 & \textbf{15.44} & \textbf{70.18} & \textbf{63.49} & \textbf{23.54} & 55.65 & 63.69 & 77.17 & 52.96\\
        InternImage Paired~\cite{wang2023internimage} & \textbf{79.01} & 36.40 & 13.29 & 70.10 & 62.02 & 16.07 & 57.62 & 65.22 & \textbf{79.27} & 53.22 \\
        InternImage Paired~\cite{wang2023internimage} + Losses + CLIP (Ours) & 77.01 & \textbf{44.69} & 11.66 & 69.86 & 62.66 & 21.95 & \textbf{62.62} & \textbf{70.23} & 74.65 & \textbf{55.04} \\
        \midrule
        ConvNeXt Adverse Only~\cite{liu2022convnet} & 72.60 & 55.54 & \textbf{18.39} & 53.52 & 60.02 & 23.83 & 42.58 & 62.52 & 43.37 & 48.04\\
        ConvNeXt Paired~\cite{liu2022convnet} & 76.27 & \textbf{58.07} & 7.54 & 59.82 & \textbf{63.39} & 17.77 & 36.29 & 63.97 & 66.08 & 49.91\\
        ConvNeXt Paired~\cite{liu2022convnet} + Losses + CLIP (Ours) & \textbf{76.30} & 47.02 & 7.42 & \textbf{73.74} & 62.39 & \textbf{24.42} & \textbf{56.95} & \textbf{71.77} & \textbf{68.69} & \textbf{54.30} \\
        \bottomrule
      \end{tabular}
  }
  \caption{\textbf{Both InternImage~\cite{wang2023internimage} and ConvNeXt~\cite{liu2022convnet} still perform as well or better on clear images when using paired-data training.}}
  \label{table:Clean-table}
\end{table*}

\subsection{Ablation Studies}
\label{sec:Ablation}
\noindent
\textbf{CLIP Injection Method}: In~\cref{table:clip-type}, we compare the performance of our model to other CLIP injection methods. We first test MultiCLIP, in which $\mathbf{T}_{\text{CLIP}}$ is concatenated in the second dimension with $\mathbf{M}_{CLIP} \in \mathbb{R}^{N \times 512}$, which is created by repeating $\vec{\mathbf{I}}_{\text{CLIP}}$ N times. The resulting matrix is $\tilde{\mathbf{W}} \in \mathbb{R}^{N \times 1024}$, which is fed into the model. This CLIP injection method lowers mIoU performance by 2.94\% compared to our proposed method. This decrease in performance can likely be attributed to not constraining the final composition of weather effects by learning a relation between $\mathbf{I}_d(x)$ and $\mathbf{T}_{\text{CLIP}}$. The other CLIP injection method is to use only a set of $N=4$ texts in the set $\mathcal{T}$ to describe the weather conditions. We find that, by limiting the size of $\mathcal{T}$, we decrease the CLIP injection layer's ability to boost model learning, leading to a lower mIoU performance by 0.64\%.


\noindent
\textbf{Feature Consistency Loss}: We compare our feature consistency loss to the rain-invariant loss~\cite{ba2022not}. We also test a step, linear, and sigmoid scheduler for the weights of our losses. Our feature consistency loss with a sigmoid schedule is shown to outperform all rain-invariant losses by at least 0.42\%, as seen in~\cref{table:FC-loss}. We find empirically that using both the ``push" and ``pull" from the rain-invariant loss leads to slightly decreased performance, which is likely due to the fact that distancing similar scenes away from each other can have the opposite effect of encouraging the model to accurately predict segmentation of novel scenes rather than focus on adaptability to adverse weather degradations. We attribute the sigmoid schedule being the best scheduler to the backbone of the model $\mathcal{F}(\cdot,\theta_1)$ undergoing general pretraining, so the model's extracted features should be close to being well conditioned to similar inputs $\textbf{I}_d(x)$ and $\textbf{J}(x)$. Therefore, we can start the feature consistency loss early to ensure the latent representation doesn't deviate too much at the start of training.

\noindent
\textbf{Output Consistency Loss}: We compare three schedulers for our output consistency loss: step, linear, and sigmoid. As seen in~\cref{table:OC-Loss}, we see that our proposed loss with a step scheduler outperforms the other schedulers. We attribute the step function being the best scheduler because the model has a random initialization of its decode head $\mathcal{G}(\cdot, \theta_2)$, meaning its semantic segmentation labels are not well conditioned to be consistent between similar inputs. Thus, we let the model train until outputs stabilize to add the output consistency loss.

\begin{table}
  \centering
  \begin{tabular}{l|c}
    \toprule
    Model & mIoU $\uparrow$ \\
    \midrule
    InternImage Paired~\cite{wang2023internimage} + MultiCLIP & 46.39\\
    InternImage Paired~\cite{wang2023internimage} + 4CLIP & 48.69\\
    InternImage Paired~\cite{wang2023internimage} + 13CLIP & \textbf{49.33}\\
    \bottomrule
  \end{tabular}
  \caption{\textbf{Ablation studies show that clip injection with 13 text prompts performs best.}}
  \label{table:clip-type}
\end{table}

\begin{table}
  \centering
  \begin{tabular}{l|c}
    \toprule
    Model & mIoU $\uparrow$\\
    \midrule
    InternImage Paired~\cite{wang2023internimage} & \textbf{45.24}\\
    ConvNeXt Paired~\cite{liu2022convnet} & 40.92 \\
    AWSS~\cite{kerim2022semantic} & 32.09\\
    \bottomrule
  \end{tabular}
  \caption{\textbf{Ablation studies show modern foundational models drastically outperform models specifically tailored for use in adverse weather conditions.} Both mIoUs are a result from evaluating their respective models on adverse weather images only.}
  \label{table:foundation}
\end{table}


\begin{table}
  \centering
  \resizebox{\columnwidth}{!}{
      \begin{tabular}{l|c}
        \toprule
        Model & mIoU $\uparrow$\\
        \midrule
        InternImage Paired~\cite{wang2023internimage} + R loss w/ Step & 48.30\\
        InternImage Paired~\cite{wang2023internimage} + R loss w/ Linear & 48.39\\
        InternImage Paired~\cite{wang2023internimage} + R loss w/ Sigmoid & 48.4\\
        InternImage Paired~\cite{wang2023internimage} + FCL w/ Step & 48.32\\
        InternImage Paired~\cite{wang2023internimage} + FCL w/ Linear & 48.63\\
        InternImage Paired~\cite{wang2023internimage} + FCL w/ Sigmoid & \textbf{48.82}\\
        \bottomrule
      \end{tabular}
  }
  \caption{\textbf{Ablation studies show that feature consistency with sigmoid scheduling performs best.}}
  \label{table:FC-loss}
\end{table}

\begin{table}
  \centering
  \resizebox{\columnwidth}{!}{
  \begin{tabular}{l|c}
    \toprule
    Model & mIoU $\uparrow$\\
    \midrule
    InternImage Paired~\cite{wang2023internimage} + OCL w/ Step & \textbf{47.98}\\
    InternImage Paired~\cite{wang2023internimage} + OCL w/ Linear & 47.43\\
    InternImage Paired~\cite{wang2023internimage} + OCL w/ Sigmoid & 47.43\\
    \bottomrule
  \end{tabular}
  }
  \caption{\textbf{Ablation studies show that our OCL with a step scheduler performs best.}}
  \label{table:OC-Loss}
\end{table}

\subsection{Model Gradient Testing}
\label{sec:exp_grad}
We validate in this section our hypothesis that foundational models prioritize learning to adapt to new scenes rather than learning to adapt to weather degradations within the same scene due to their respective effects on minimizing the cross entropy loss. We do so by taking the pretrained InternImage model~\cite{wang2023internimage}, and training it on clear images that were previously unseen to the model. Then, we compute the loss on the images capturing the same scene, but under adverse weather instead. As these models are trained using gradient-based optimization, we measure the L2-norm of the gradient by backpropagating through the weights to quantify the change induced by samples, $||G_d||_2$. Additionally, we compute the same loss on an image that captures a novel scene (disjoint from those in the training set) under clear weather, again measuring the L2-norm of the gradient $||G_c||_2$. This was repeated for 78 trials and the L2-norms denote their means. On average, we see that $||G_d||_2 = 40.26$ and $||G_c||_2 = 58.21$. Given these results, we find that giving the model a new scene to learn leads to larger weight updates than does existing scenes observed by the model but under adverse conditions, showing that these models will prioritize learning the former rather than the latter to minimize cross entropy loss. As mentioned before, this is likely due to the fact that inter-scene variance is much higher as a result of numerous factors such as scene environment, structure, image resolution, camera parameters, etc.
\section{Conclusion}

In this paper, we investigate the performance gap of foundational models on the semantic segmentation task in adverse weather conditions. We introduced a paired-training method that improves mIOU by up to 18.4\%. We do this by training on finely paired clear and adverse weather images which have the same underlying semantic scene labels. This allows the network to differentiate between learning the semantic labels of new scenes and learning to maintain accuracy despite weather degradations. To ``balance out" the fact that focusing on the former better minimizes cross entropy loss, we employ two consistency losses (FCL and OCL), as well as a language-guided CLIP Injection Layer, to focus the model on the latter.

This work represents an initial attempt at improving the performance of foundational models in adverse weather conditions through the use of a paired dataset training paradigm. We hope that by introducing our \dname, we inspire further research into the benefits of training on clear and adverse weather image pairs, with the eventual hope that future models can achieve ADE-20K or Cityscapes levels of accuracy despite the presence of adverse visual degradations such as weather. 

{
    \small
    \bibliographystyle{ieeenat_fullname}
    \bibliography{main}

\begin{thebibliography}{48}
\providecommand{\natexlab}[1]{#1}
\providecommand{\url}[1]{\texttt{#1}}
\expandafter\ifx\csname urlstyle\endcsname\relax
  \providecommand{\doi}[1]{doi: #1}\else
  \providecommand{\doi}{doi: \begingroup \urlstyle{rm}\Url}\fi

\bibitem[Ba et~al.(2022)Ba, Zhang, Yang, Suzuki, Pfahnl, Chandrappa, de~Melo, You, Soatto, Wong, et~al.]{ba2022not}
Yunhao Ba, Howard Zhang, Ethan Yang, Akira Suzuki, Arnold Pfahnl, Chethan~Chinder Chandrappa, Celso~M de Melo, Suya You, Stefano Soatto, Alex Wong, et~al.
\newblock Not just streaks: Towards ground truth for single image deraining.
\newblock In \emph{European Conference on Computer Vision}, pages 723--740. Springer, 2022.

\bibitem[Chen et~al.(2020)Chen, Fang, Ding, Tsai, and Kuo]{chen2020jstasr}
Wei-Ting Chen, Hao-Yu Fang, Jian-Jiun Ding, Cheng-Che Tsai, and Sy-Yen Kuo.
\newblock Jstasr: Joint size and transparency-aware snow removal algorithm based on modified partial convolution and veiling effect removal.
\newblock In \emph{Computer Vision--ECCV 2020: 16th European Conference, Glasgow, UK, August 23--28, 2020, Proceedings, Part XXI 16}, pages 754--770. Springer, 2020.

\bibitem[Chen et~al.(2021)Chen, Fang, Hsieh, Tsai, Chen, Ding, Kuo, et~al.]{chen2021all}
Wei-Ting Chen, Hao-Yu Fang, Cheng-Lin Hsieh, Cheng-Che Tsai, I Chen, Jian-Jiun Ding, Sy-Yen Kuo, et~al.
\newblock All snow removed: Single image desnowing algorithm using hierarchical dual-tree complex wavelet representation and contradict channel loss.
\newblock In \emph{Proceedings of the IEEE/CVF International Conference on Computer Vision}, pages 4196--4205, 2021.

\bibitem[Contributors(2020)]{mmseg2020}
MMSegmentation Contributors.
\newblock {MMSegmentation}: Openmmlab semantic segmentation toolbox and benchmark.
\newblock \url{https://github.com/open-mmlab/mmsegmentation}, 2020.

\bibitem[Cordts et~al.(2016)Cordts, Omran, Ramos, Rehfeld, Enzweiler, Benenson, Franke, Roth, and Schiele]{cordts2016cityscapes}
Marius Cordts, Mohamed Omran, Sebastian Ramos, Timo Rehfeld, Markus Enzweiler, Rodrigo Benenson, Uwe Franke, Stefan Roth, and Bernt Schiele.
\newblock The cityscapes dataset for semantic urban scene understanding.
\newblock In \emph{Proceedings of the IEEE conference on computer vision and pattern recognition}, pages 3213--3223, 2016.

\bibitem[Deng et~al.(2018)Deng, Huang, Zhao, and Jiang]{deng2018directional}
Liang-Jian Deng, Ting-Zhu Huang, Xi-Le Zhao, and Tai-Xiang Jiang.
\newblock A directional global sparse model for single image rain removal.
\newblock \emph{Applied Mathematical Modelling}, 59:\penalty0 662--679, 2018.

\bibitem[Ess et~al.(2009)Ess, M{\"u}ller, Grabner, and Van~Gool]{ess2009segmentation}
Andreas Ess, Tobias M{\"u}ller, Helmut Grabner, and Luc Van~Gool.
\newblock Segmentation-based urban traffic scene understanding.
\newblock In \emph{BMVC}, page~2. Citeseer, 2009.

\bibitem[Fu et~al.(2017)Fu, Huang, Zeng, Huang, Ding, and Paisley]{fu2017removing}
Xueyang Fu, Jiabin Huang, Delu Zeng, Yue Huang, Xinghao Ding, and John Paisley.
\newblock Removing rain from single images via a deep detail network.
\newblock In \emph{Proceedings of the IEEE conference on computer vision and pattern recognition}, pages 3855--3863, 2017.

\bibitem[Gupta et~al.(2015)Gupta, Arbel{\'a}ez, Girshick, and Malik]{gupta2015indoor}
Saurabh Gupta, Pablo Arbel{\'a}ez, Ross Girshick, and Jitendra Malik.
\newblock Indoor scene understanding with rgb-d images: Bottom-up segmentation, object detection and semantic segmentation.
\newblock \emph{International Journal of Computer Vision}, 112:\penalty0 133--149, 2015.

\bibitem[He et~al.(2010)He, Sun, and Tang]{he2010single}
Kaiming He, Jian Sun, and Xiaoou Tang.
\newblock Single image haze removal using dark channel prior.
\newblock \emph{IEEE transactions on pattern analysis and machine intelligence}, 33\penalty0 (12):\penalty0 2341--2353, 2010.

\bibitem[Hong et~al.(2015)Hong, Noh, and Han]{hong2015decoupled}
Seunghoon Hong, Hyeonwoo Noh, and Bohyung Han.
\newblock Decoupled deep neural network for semi-supervised semantic segmentation.
\newblock \emph{Advances in neural information processing systems}, 28, 2015.

\bibitem[Hu et~al.(2021)Hu, Cao, Lu, Zhang, Wang, Li, Huang, Shao, and Ji]{hu2021istr}
Jie Hu, Liujuan Cao, Yao Lu, ShengChuan Zhang, Yan Wang, Ke Li, Feiyue Huang, Ling Shao, and Rongrong Ji.
\newblock Istr: End-to-end instance segmentation with transformers.
\newblock \emph{arXiv preprint arXiv:2105.00637}, 2021.

\bibitem[Hu et~al.(2018)Hu, Dollár, He, Darrell, and Girshick]{hu2018learning}
Ronghang Hu, Piotr Dollár, Kaiming He, Trevor Darrell, and Ross Girshick.
\newblock Learning to segment every thing, 2018.

\bibitem[Kerim et~al.(2022)Kerim, Chamone, Ramos, Marcolino, Nascimento, and Jiang]{kerim2022semantic}
Abdulrahman Kerim, Felipe Chamone, Washington Ramos, Leandro~Soriano Marcolino, Erickson~R Nascimento, and Richard Jiang.
\newblock Semantic segmentation under adverse conditions: a weather and nighttime-aware synthetic data-based approach.
\newblock \emph{arXiv preprint arXiv:2210.05626}, 2022.

\bibitem[Kim and Seok(2018)]{kim2018indoor}
Wonsuk Kim and Junhee Seok.
\newblock Indoor semantic segmentation for robot navigating on mobile.
\newblock In \emph{2018 Tenth International Conference on Ubiquitous and Future Networks (ICUFN)}, pages 22--25. IEEE, 2018.

\bibitem[Li et~al.(2018)Li, He, Zhang, Chang, Dong, and Lin]{li2018non}
Guanbin Li, Xiang He, Wei Zhang, Huiyou Chang, Le Dong, and Liang Lin.
\newblock Non-locally enhanced encoder-decoder network for single image de-raining.
\newblock In \emph{Proceedings of the 26th ACM international conference on Multimedia}, pages 1056--1064, 2018.

\bibitem[Li et~al.(2022)Li, Li, Xiong, and Hoi]{li2022blip}
Junnan Li, Dongxu Li, Caiming Xiong, and Steven Hoi.
\newblock Blip: Bootstrapping language-image pre-training for unified vision-language understanding and generation.
\newblock In \emph{International Conference on Machine Learning}, pages 12888--12900. PMLR, 2022.

\bibitem[Li et~al.(2009)Li, Socher, and Fei-Fei]{li2009towards}
Li-Jia Li, Richard Socher, and Li Fei-Fei.
\newblock Towards total scene understanding: Classification, annotation and segmentation in an automatic framework.
\newblock In \emph{2009 IEEE Conference on Computer Vision and Pattern Recognition}, pages 2036--2043. IEEE, 2009.

\bibitem[Li et~al.(2019)Li, Cheong, and Tan]{li2019heavy}
Ruoteng Li, Loong-Fah Cheong, and Robby~T Tan.
\newblock Heavy rain image restoration: Integrating physics model and conditional adversarial learning.
\newblock In \emph{Proceedings of the IEEE/CVF conference on computer vision and pattern recognition}, pages 1633--1642, 2019.

\bibitem[Li et~al.(2020)Li, Tan, and Cheong]{li2020all}
Ruoteng Li, Robby~T Tan, and Loong-Fah Cheong.
\newblock All in one bad weather removal using architectural search.
\newblock In \emph{Proceedings of the IEEE/CVF conference on computer vision and pattern recognition}, pages 3175--3185, 2020.

\bibitem[Li et~al.(2016)Li, Tan, Guo, Lu, and Brown]{li2016rain}
Yu Li, Robby~T Tan, Xiaojie Guo, Jiangbo Lu, and Michael~S Brown.
\newblock Rain streak removal using layer priors.
\newblock In \emph{Proceedings of the IEEE conference on computer vision and pattern recognition}, pages 2736--2744, 2016.

\bibitem[Lin et~al.(2017)Lin, Milan, Shen, and Reid]{lin2017refinenet}
Guosheng Lin, Anton Milan, Chunhua Shen, and Ian Reid.
\newblock Refinenet: Multi-path refinement networks for high-resolution semantic segmentation.
\newblock In \emph{Proceedings of the IEEE conference on computer vision and pattern recognition}, pages 1925--1934, 2017.

\bibitem[Liu et~al.(2018)Liu, Jaw, Huang, and Hwang]{liu2018desnownet}
Yun-Fu Liu, Da-Wei Jaw, Shih-Chia Huang, and Jenq-Neng Hwang.
\newblock Desnownet: Context-aware deep network for snow removal.
\newblock \emph{IEEE Transactions on Image Processing}, 27\penalty0 (6):\penalty0 3064--3073, 2018.

\bibitem[Liu et~al.(2021)Liu, Lin, Cao, Hu, Wei, Zhang, Lin, and Guo]{liu2021swin}
Ze Liu, Yutong Lin, Yue Cao, Han Hu, Yixuan Wei, Zheng Zhang, Stephen Lin, and Baining Guo.
\newblock Swin transformer: Hierarchical vision transformer using shifted windows.
\newblock In \emph{Proceedings of the IEEE/CVF international conference on computer vision}, pages 10012--10022, 2021.

\bibitem[Liu et~al.(2022{\natexlab{a}})Liu, Hu, Lin, Yao, Xie, Wei, Ning, Cao, Zhang, Dong, et~al.]{liu2022swin}
Ze Liu, Han Hu, Yutong Lin, Zhuliang Yao, Zhenda Xie, Yixuan Wei, Jia Ning, Yue Cao, Zheng Zhang, Li Dong, et~al.
\newblock Swin transformer v2: Scaling up capacity and resolution.
\newblock In \emph{Proceedings of the IEEE/CVF conference on computer vision and pattern recognition}, pages 12009--12019, 2022{\natexlab{a}}.

\bibitem[Liu et~al.(2022{\natexlab{b}})Liu, Mao, Wu, Feichtenhofer, Darrell, and Xie]{liu2022convnet}
Zhuang Liu, Hanzi Mao, Chao-Yuan Wu, Christoph Feichtenhofer, Trevor Darrell, and Saining Xie.
\newblock A convnet for the 2020s.
\newblock In \emph{Proceedings of the IEEE/CVF conference on computer vision and pattern recognition}, pages 11976--11986, 2022{\natexlab{b}}.

\bibitem[Long et~al.(2015)Long, Shelhamer, and Darrell]{long2015fully}
Jonathan Long, Evan Shelhamer, and Trevor Darrell.
\newblock Fully convolutional networks for semantic segmentation.
\newblock In \emph{Proceedings of the IEEE conference on computer vision and pattern recognition}, pages 3431--3440, 2015.

\bibitem[Milioto and Stachniss(2019)]{milioto2019bonnet}
Andres Milioto and Cyrill Stachniss.
\newblock Bonnet: An open-source training and deployment framework for semantic segmentation in robotics using cnns.
\newblock In \emph{2019 international conference on robotics and automation (ICRA)}, pages 7094--7100. IEEE, 2019.

\bibitem[Milioto et~al.(2018)Milioto, Lottes, and Stachniss]{milioto2018real}
Andres Milioto, Philipp Lottes, and Cyrill Stachniss.
\newblock Real-time semantic segmentation of crop and weed for precision agriculture robots leveraging background knowledge in cnns.
\newblock In \emph{2018 IEEE international conference on robotics and automation (ICRA)}, pages 2229--2235. IEEE, 2018.

\bibitem[Nekrasov et~al.(2019)Nekrasov, Dharmasiri, Spek, Drummond, Shen, and Reid]{nekrasov2019real}
Vladimir Nekrasov, Thanuja Dharmasiri, Andrew Spek, Tom Drummond, Chunhua Shen, and Ian Reid.
\newblock Real-time joint semantic segmentation and depth estimation using asymmetric annotations.
\newblock In \emph{2019 International Conference on Robotics and Automation (ICRA)}, pages 7101--7107. IEEE, 2019.

\bibitem[Nichol et~al.(2021)Nichol, Dhariwal, Ramesh, Shyam, Mishkin, McGrew, Sutskever, and Chen]{nichol2021glide}
Alex Nichol, Prafulla Dhariwal, Aditya Ramesh, Pranav Shyam, Pamela Mishkin, Bob McGrew, Ilya Sutskever, and Mark Chen.
\newblock Glide: Towards photorealistic image generation and editing with text-guided diffusion models.
\newblock \emph{arXiv preprint arXiv:2112.10741}, 2021.

\bibitem[Radford et~al.(2021)Radford, Kim, Hallacy, Ramesh, Goh, Agarwal, Sastry, Askell, Mishkin, Clark, et~al.]{radford2021learning}
Alec Radford, Jong~Wook Kim, Chris Hallacy, Aditya Ramesh, Gabriel Goh, Sandhini Agarwal, Girish Sastry, Amanda Askell, Pamela Mishkin, Jack Clark, et~al.
\newblock Learning transferable visual models from natural language supervision.
\newblock In \emph{International conference on machine learning}, pages 8748--8763. PMLR, 2021.

\bibitem[Rombach et~al.(2022)Rombach, Blattmann, Lorenz, Esser, and Ommer]{rombach2022high}
Robin Rombach, Andreas Blattmann, Dominik Lorenz, Patrick Esser, and Bj{\"o}rn Ommer.
\newblock High-resolution image synthesis with latent diffusion models.
\newblock In \emph{Proceedings of the IEEE/CVF conference on computer vision and pattern recognition}, pages 10684--10695, 2022.

\bibitem[Sakaridis et~al.(2018)Sakaridis, Dai, and Van~Gool]{sakaridis2018semantic}
Christos Sakaridis, Dengxin Dai, and Luc Van~Gool.
\newblock Semantic foggy scene understanding with synthetic data.
\newblock \emph{International Journal of Computer Vision}, 126:\penalty0 973--992, 2018.

\bibitem[Sakaridis et~al.(2021)Sakaridis, Dai, and Van~Gool]{sakaridis2021acdc}
Christos Sakaridis, Dengxin Dai, and Luc Van~Gool.
\newblock Acdc: The adverse conditions dataset with correspondences for semantic driving scene understanding.
\newblock In \emph{Proceedings of the IEEE/CVF International Conference on Computer Vision}, pages 10765--10775, 2021.

\bibitem[Siam et~al.(2018)Siam, Gamal, Abdel-Razek, Yogamani, and Jagersand]{siam2018rtseg}
Mennatullah Siam, Mostafa Gamal, Moemen Abdel-Razek, Senthil Yogamani, and Martin Jagersand.
\newblock Rtseg: Real-time semantic segmentation comparative study.
\newblock In \emph{2018 25th IEEE International Conference on Image Processing (ICIP)}, pages 1603--1607. IEEE, 2018.

\bibitem[Tan(2008)]{tan2008visibility}
Robby~T Tan.
\newblock Visibility in bad weather from a single image.
\newblock In \emph{2008 IEEE conference on computer vision and pattern recognition}, pages 1--8. IEEE, 2008.

\bibitem[Valanarasu et~al.(2022)Valanarasu, Yasarla, and Patel]{valanarasu2022transweather}
Jeya Maria~Jose Valanarasu, Rajeev Yasarla, and Vishal~M Patel.
\newblock Transweather: Transformer-based restoration of images degraded by adverse weather conditions.
\newblock In \emph{Proceedings of the IEEE/CVF Conference on Computer Vision and Pattern Recognition}, pages 2353--2363, 2022.

\bibitem[Wang et~al.(2020)Wang, Xie, Zhao, and Meng]{wang2020model}
Hong Wang, Qi Xie, Qian Zhao, and Deyu Meng.
\newblock A model-driven deep neural network for single image rain removal.
\newblock In \emph{Proceedings of the IEEE/CVF conference on computer vision and pattern recognition}, pages 3103--3112, 2020.

\bibitem[Wang et~al.(2019)Wang, Yang, Xu, Chen, Zhang, and Lau]{wang2019spatial}
Tianyu Wang, Xin Yang, Ke Xu, Shaozhe Chen, Qiang Zhang, and Rynson~WH Lau.
\newblock Spatial attentive single-image deraining with a high quality real rain dataset.
\newblock In \emph{Proceedings of the IEEE/CVF Conference on Computer Vision and Pattern Recognition}, pages 12270--12279, 2019.

\bibitem[Wang et~al.(2023)Wang, Dai, Chen, Huang, Li, Zhu, Hu, Lu, Lu, Li, et~al.]{wang2023internimage}
Wenhai Wang, Jifeng Dai, Zhe Chen, Zhenhang Huang, Zhiqi Li, Xizhou Zhu, Xiaowei Hu, Tong Lu, Lewei Lu, Hongsheng Li, et~al.
\newblock Internimage: Exploring large-scale vision foundation models with deformable convolutions.
\newblock In \emph{Proceedings of the IEEE/CVF Conference on Computer Vision and Pattern Recognition}, pages 14408--14419, 2023.

\bibitem[Xie et~al.(2021)Xie, Wang, Yu, Anandkumar, Alvarez, and Luo]{xie2021segformer}
Enze Xie, Wenhai Wang, Zhiding Yu, Anima Anandkumar, Jose~M Alvarez, and Ping Luo.
\newblock Segformer: Simple and efficient design for semantic segmentation with transformers.
\newblock \emph{Advances in Neural Information Processing Systems}, 34:\penalty0 12077--12090, 2021.

\bibitem[Yasarla and Patel(2019)]{yasarla2019uncertainty}
Rajeev Yasarla and Vishal~M Patel.
\newblock Uncertainty guided multi-scale residual learning-using a cycle spinning cnn for single image de-raining.
\newblock In \emph{Proceedings of the IEEE/CVF conference on computer vision and pattern recognition}, pages 8405--8414, 2019.

\bibitem[Zhang and Patel(2018)]{zhang2018density}
He Zhang and Vishal~M Patel.
\newblock Density-aware single image de-raining using a multi-stream dense network.
\newblock In \emph{Proceedings of the IEEE conference on computer vision and pattern recognition}, pages 695--704, 2018.

\bibitem[Zhang et~al.(2023)Zhang, Ba, Yang, Mehra, Gella, Suzuki, Pfahnl, Chandrappa, Wong, and Kadambi]{zhang2023weatherstream}
Howard Zhang, Yunhao Ba, Ethan Yang, Varan Mehra, Blake Gella, Akira Suzuki, Arnold Pfahnl, Chethan~Chinder Chandrappa, Alex Wong, and Achuta Kadambi.
\newblock Weatherstream: Light transport automation of single image deweathering.
\newblock In \emph{Proceedings of the IEEE/CVF Conference on Computer Vision and Pattern Recognition}, pages 13499--13509, 2023.

\bibitem[Zhao et~al.(2018)Zhao, Qi, Shen, Shi, and Jia]{zhao2018icnet}
Hengshuang Zhao, Xiaojuan Qi, Xiaoyong Shen, Jianping Shi, and Jiaya Jia.
\newblock Icnet for real-time semantic segmentation on high-resolution images.
\newblock In \emph{Proceedings of the European conference on computer vision (ECCV)}, pages 405--420, 2018.

\bibitem[Zhou et~al.(2017)Zhou, Zhao, Puig, Fidler, Barriuso, and Torralba]{zhou2017scene}
Bolei Zhou, Hang Zhao, Xavier Puig, Sanja Fidler, Adela Barriuso, and Antonio Torralba.
\newblock Scene parsing through ade20k dataset.
\newblock In \emph{Proceedings of the IEEE conference on computer vision and pattern recognition}, pages 633--641, 2017.

\bibitem[Zhu et~al.(2017)Zhu, Fu, Lischinski, and Heng]{zhu2017joint}
Lei Zhu, Chi-Wing Fu, Dani Lischinski, and Pheng-Ann Heng.
\newblock Joint bi-layer optimization for single-image rain streak removal.
\newblock In \emph{Proceedings of the IEEE international conference on computer vision}, pages 2526--2534, 2017.

\end{thebibliography}
}

\renewcommand\thesection{\Alph{section}} 
\renewcommand\thesubsection{\thesection.\alph{subsection}} 
\renewcommand\thefigure{\Alph{figure}} 
\renewcommand\thetable{\Alph{table}} 
\setcounter{figure}{0}
\setcounter{section}{0}
\setcounter{table}{0}



\onecolumn



\clearpage
\section*{Supplementary Contents}

This supplement is organized as follows:
\begin{itemize}
    \item \Cref{sec:acdc} shows some comparison metrics on the ACDC dataset.
    \item \Cref{sec:qual} shows some qualitative results from the \dname\ test set.
    \item \Cref{sec:quant} shows some comparison metrics from the \dname\ test set on additional foundational models.
    \item \Cref{sec:implementation} shows the implementation details of models.
    \item \Cref{sec:failure} shows some failure modes for our training method.
\end{itemize}
\vspace{5pt}
\noindent
Our \dname\ will be released conditional on acceptance.

\section{Results on ACDC Dataset}
\label{sec:acdc}
In \Cref{tab:ACDC-table}, we train and evaluate InternImage~\cite{wang2023internimage} and our augmented InternImage on the ACDC dataset~\cite{sakaridis2021acdc}. We also finetune on the ACDC dataset with two of the InternImage models trained on the \dname, one with CLIP~\cite{radford2021learning} and losses and the other base. As shown, pretraining the model on our dataset helps performance significantly. Moreover, the model with CLIP benefits more than the base model from pretraining, which we attribute to the model being exposed to clear and adverse weather conditions as well as the consistency losses. We also see observe that all CLIP models perform better than their base counter parts. As discussed in \cref{sec:related_adverse} in the main paper, the ACDC dataset does not have accurate clean and adverse weather image pairs. Thus, we do not take full advantage of CLIP nor do we train with our consistency losses, leading to a smaller increase in performance compared to the performance increase when training on our \dname.

As discussed in \cref{sec:results} in the main paper, foundational models are able to significantly outperform single-task semantic segmentation models that are specifically designed to perform well on adverse weather conditions. This is also shown in \Cref{tab:ACDC-table}, as the AWSS model~\cite{kerim2022semantic}, which was trained on CityScapes~\cite{cordts2016cityscapes} and ACDC, performs significantly worse than InternImage, which was only trained on ACDC. 

\begin{table*}[h]
  \centering
  \resizebox{\textwidth}{!}{
      \begin{tabular}{ccccccccccccccccccccc}
        \toprule
        Model & \rotatebox{90}{Road} & \rotatebox{90}{Sidewalk} & \rotatebox{90}{Building} & \rotatebox{90}{Wall} & \rotatebox{90}{Fence} & \rotatebox{90}{Pole} & \rotatebox{90}{Traffic Light} & \rotatebox{90}{Traffic Sign} & \rotatebox{90}{Vegetation} & \rotatebox{90}{Terrain} & \rotatebox{90}{Sky} & \rotatebox{90}{Person} & \rotatebox{90}{Rider} & \rotatebox{90}{Car} & \rotatebox{90}{Truck} & \rotatebox{90}{Bus} & \rotatebox{90}{Train} & \rotatebox{90}{Motorcycle} & \rotatebox{90}{Bicycle} & mIoU $\uparrow$ \\
        \midrule
        AWSS~\cite{kerim2022semantic} & 79 & 40 & 63 & - & - & 25 & 26 & 33 & 69 & - & 66 & 32 & - & 52 & - & - & - & - & - & 49\\
        \midrule
        InternImage ~\cite{wang2023internimage} & 
94.36 & 77.18 & 86.87 & 58.24 & 53.69 & 54.94 & 66.05 & 57.82 & 86.23 & 51.99 & 95.38 & 54.69 & 19.69 & 82.68 & 77.29 & 90.73 & 90.23 & 36.84 & 49.12 & 67.58 \\
        InternImage ~\cite{wang2023internimage} Pretrained & \textbf{94.45} & \textbf{77.19} & 87.08 & 60.28 & \textbf{55.72} & 55.37 & 65.62 & \textbf{58.89} & 86.35 & 50.89 & \textbf{95.58} & 54.60 & 17.65 & \textbf{83.66} & 79.37 & 90.83 & 90.14 & 36.66 & 45.65 & 67.68 \\
        InternImage ~\cite{wang2023internimage} + CLIP (Ours) & 94.35 & 77.17 & \textbf{87.16} & \textbf{60.60} & 54.72 & 55.71 & 66.48 & 57.56 & 86.31 & \textbf{52.51} & 95.44 & \textbf{55.39} & 14.87 & 82.88 & 78.26 & \textbf{91.49} & \textbf{90.25} & 37.61 & \textbf{49.27} & 67.79\\
        InternImage ~\cite{wang2023internimage} Pretrained + CLIP (Ours) & 94.19 & 76.78 & 87.13 & 59.82 & 54.54 & \textbf{55.78} & \textbf{66.50} & 57.82 & \textbf{86.36} & 52.08 & 95.55 & 54.47 & \textbf{23.02} & 83.45 & \textbf{79.45} & 90.66 & 89.94 & \textbf{40.96} & 47.79 & \textbf{68.22}\\
        \midrule
      \end{tabular}
    }
  \caption{\textbf{On InternImage~\cite{wang2023internimage}, our proposed CLIP method outperforms the base model and pretraining it on our \dname\ always boosts mIoU.} In addition, foundational models significantly outperform AWSS~\cite{kerim2022semantic}, our comparison for a single-task adverse weather semantic segmentation model. For the AWSS model, the "-" means that they did not provide the IoU for this class.}
  \label{tab:ACDC-table}
\end{table*}

\section{Additional Qualitative Results on \dname}
\label{sec:qual}
\begin{figure*}[t]
    \centering
    \includegraphics [width=\linewidth]{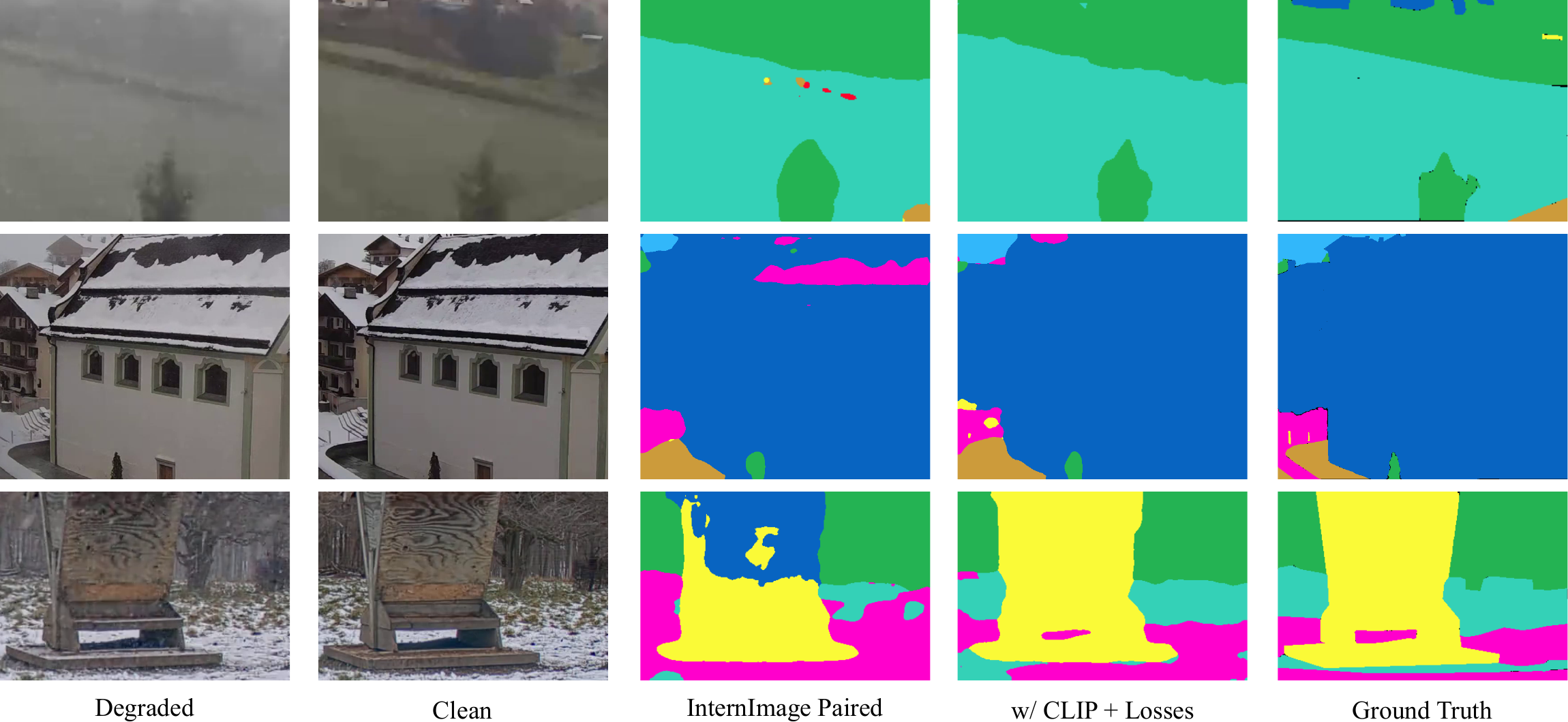}
    \caption{\textbf{Qualitative results from the InternImage trained with only paired images and InternImage trained with our proposed CLIP guidance and consistency losses.} The degraded image, clean image, and ground truth semantic segmentation maps are also included for reference. As discussed in~\Cref{sec:implementation}, MMsegmentation's decoders produce a segmentation map that is 0.25 the size of the original image and use billinear interpolation to get the full map. Thus, both models often lack the ability to produce high frequency details and have much smoother labels.}
    \label{fig:qualititative}
\end{figure*}

In~\Cref{fig:qualititative}, we show the qualitative results of our models compared to the paired only model. As discussed in~\Cref{sec:implementation}, MMSegmentation~\cite{mmseg2020} upscales by 4, meaning many high frequency details are not present in the outputs.

\section{Additional Foundational Model Tests}
\label{sec:quant}
In ~\cref{tab:add-quant-degraded}, we include the results of the Swin Transformer model~\cite{liu2021swin} when trained using our paired training method on our \dname\ with CLIP and evaluated on degraded images only. We also include the other results from \cref{fig:model} in the main paper for comparisons. While the base version and our augmented version of the Swin Transformer model perform well, the degraded only version performs much worse than ours, the paired version, ConvNeXt~\cite{liu2022convnet} adverse only, or InternImage~\cite{wang2023internimage} adverse only. We attribute this to Swin having a transformer backbone, compared to ConvNeXt's convolutional and InternImage's deformable convolutional backbones. As transforms in general require more data to become robust than convolutional networks, we believe Swin suffers more than the other models when clean images are removed.

\begin{table*}
  \centering
  \resizebox{\textwidth}{!}{
      \begin{tabular}{ccccccccccc}
        \toprule
        Model & \rotatebox{90}{Tree} & \rotatebox{90}{Struc.} & \rotatebox{90}{Road} & \rotatebox{90}{T-Snow} & \rotatebox{90}{T-Veg.} & \rotatebox{90}{T-Other} & \rotatebox{90}{Stone} & \rotatebox{90}{Building} & \rotatebox{90}{Sky} & mIoU $\uparrow$ \\
        \midrule
        InternImage Adverse Only~\cite{wang2023internimage} & 71.78 & 41.01 & \textbf{10.20} & 64.95 & 60.40 & \textbf{22.96} & 17.28 & 64.84 & 36.45 & 43.32 \\
        InternImage Paired~\cite{wang2023internimage}& 71.23 & 35.26 & 6.73 & \textbf{66.72} & 59.97 & 16.61 & 34.0 & 67.87 & 48.74 & 45.24\\
        InternImage Paired~\cite{wang2023internimage} + Losses + CLIP (Ours) & \textbf{74.73} & \textbf{46.71} & 6.60 & 66.55 & \textbf{64.8} & 19.33 & \textbf{50.07} & \textbf{73.99} & \textbf{58.98} & \textbf{51.31}\\
        \midrule
        ConvNeXt Adverse Only~\cite{liu2022convnet} & 66.40 & 45.83 & 7.66 & 45.43 & \textbf{58.67} & 14.62 & 24.69 & 59.45 & 37.88 & 40.07\\
        ConvNeXt Paired~\cite{liu2022convnet} & 62.32 & \textbf{53.34} & 5.20 & 51.14 & 53.70 & \textbf{16.33} & 20.69 & 60.69 & \textbf{44.69} & 40.92\\
        ConvNeXt Paired~\cite{liu2022convnet} + Losses + CLIP (Ours) & \textbf{68.74} & 39.63 & \textbf{7.80} & \textbf{56.10} & 57.77 & 15.21 & \textbf{40.41} & \textbf{69.33} & 40.26 & \textbf{43.92}\\
        \midrule
        Swin Adverse Only~\cite{liu2021swin} &  63.53 & 29.93 & 2.93 & 48.42 & 61.36 & 2.88 & 0.04 & 59.28 & \textbf{64.01} & 33.24\\
        Swin Paired~\cite{liu2021swin} & \textbf{74.03} & \textbf{32.07} & 4.56 & 55.10 & \textbf{62.09} & 16.19 & \textbf{45.00} & 64.50 & 35.92 & 43.72\\
        Swin Paired~\cite{liu2021swin} + Losses + CLIP (Ours) & 68.17 & 31.43 & \textbf{29.29} & \textbf{64.90} & 53.84 & \textbf{20.43} & 43.46 & \textbf{67.19} & 45.04 & \textbf{47.08}\\
        \bottomrule
      \end{tabular}
    }
  \caption{\textbf{Our proposed paired training method outperforms standard fine-tuning on adverse images only for both InternImage~\cite{wang2023internimage}, ConvNeXt~\cite{liu2022convnet}, and Swin Transformer~\cite{liu2021swin}.} We show that our models outperform their adverse only and paired only counter parts. We also show that our method is generalizable to both convolutional and transformer model architectures.}
  \label{tab:add-quant-degraded}
\end{table*} 

In ~\cref{tab:add-quant-clean}, we also include the results of the Swin Transformer model when trained using our paired training method on our \dname\ and evaluated on clean images only. We observe that the Swin Transformer model trained on only adverse images performs much worse compared to the Swin Transformer model trained on both clean and adverse weather images. This decrease in performance is much greater compared to ConvNeXt and InternImage. Again, we attribute this to Swin being a transformer model, which makes it hard for the model to perform outside its training domain. We also observe that the adverse only model decreases in performance when testing on clean images instead of adverse weather images, which we also attribute to Swin being a transformer architecture.

\begin{table*}
  \centering
  \resizebox{\textwidth}{!}{
      \begin{tabular}{ccccccccccc}
        \toprule
        Model & \rotatebox{90}{Tree} & \rotatebox{90}{Struc.} & \rotatebox{90}{Road} & \rotatebox{90}{T-Snow} & \rotatebox{90}{T-Veg.} & \rotatebox{90}{T-Other} & \rotatebox{90}{Stone} & \rotatebox{90}{Building} & \rotatebox{90}{Sky} & mIoU $\uparrow$\\
        \midrule
        InternImage Adverse Only~\cite{wang2023internimage} & 77.32 & 30.18 & \textbf{15.44} & \textbf{70.18} & \textbf{63.49} & \textbf{23.54} & 55.65 & 63.69 & 77.17 & 52.96\\
        InternImage Paired~\cite{wang2023internimage} & \textbf{79.01} & 36.40 & 13.29 & 70.10 & 62.02 & 16.07 & 57.62 & 65.22 & \textbf{79.27} & 53.22 \\
        InternImage Paired~\cite{wang2023internimage} + Losses + CLIP (Ours) & 77.01 & \textbf{44.69} & 11.66 & 69.86 & 62.66 & 21.95 & \textbf{62.62} & \textbf{70.23} & 74.65 & \textbf{55.04} \\
        \midrule
        ConvNeXt Adverse Only~\cite{liu2022convnet} & 72.60 & 55.54 & \textbf{18.39} & 53.52 & 60.02 & 23.83 & 42.58 & 62.52 & 43.37 & 48.04\\
        ConvNeXt Paired~\cite{liu2022convnet} & 76.27 & \textbf{58.07} & 7.54 & 59.82 & \textbf{63.39} & 17.77 & 36.29 & 63.97 & 66.08 & 49.91\\
        ConvNeXt Paired~\cite{liu2022convnet} + Losses + CLIP (Ours) & \textbf{76.30} & 47.02 & 7.42 & \textbf{73.74} & 62.39 & \textbf{24.42} & \textbf{56.95} & \textbf{71.77} & \textbf{68.69} & \textbf{54.30} \\
        \midrule
        Swin Adverse Only~\cite{liu2021swin} & 66.63 & 37.81 & 10.44 & 45.44 & 62.55 & 0.00 & 0.96 & 53.72 & 32.63 & 31.62\\
        Swin Paired~\cite{liu2021swin} & \textbf{78.63} & \textbf{39.04} & 7.88 & 61.03 & \textbf{62.90} & 8.94 & \textbf{55.14} & 68.92 & 70.42 & 50.32\\
        Swin Paired~\cite{liu2021swin} + Losses + CLIP (Ours) &  71.07 & 36.63 & \textbf{24.81} & \textbf{66.72} & 56.53 & \textbf{26.74} & 53.43 & \textbf{69.17} & \textbf{73.31} & \textbf{53.16}\\
        \bottomrule
      \end{tabular}
  }
  \caption{\textbf{Both InternImage~\cite{wang2023internimage}, ConvNeXt~\cite{liu2022convnet}, and Swin Transformer~\cite{liu2021swin} still perform as well or better on clear images when using paired-data training.}}
  \label{tab:add-quant-clean}
\end{table*}

\section{Implementation Details}
\label{sec:implementation}
All foundational models use their official implementation or MMSegmentation~\cite{mmseg2020} implementation. Moreover, their hyperparameters and optimizers were also retained. Due to the limitations of CLIP's~\cite{radford2021learning} input and WeatherStream~\cite{zhang2023weatherstream}, we used crop sizes of 224$\times$224 during training on \dname. Most decode heads from MMSegmentation only predict a segmentation map that is $0.25$ the original image's size and use billinear interpolation to scale it to the full size. Thus, most predictions in the output are smooth. For our models trained on the ACDC dataset~\cite{sakaridis2021acdc}, we use a 448$\times$448 crop during training and interpolate to 224$\times$224 for CLIP. We do this due to the higher resolution of the ACDC dataset, which requires a larger crop to get a better understanding of the scene. Each model was trained on a single NVIDIA 3090 GPU. For our CLIP augmented models, we use the CLIP VIT-B/32 model. We chose this model to fit to the WeatherStream dataset and be compact enough to store with the semantic segmentation on a single GPU.

\subsection{InternImage}
We use InternImage's config for their largest model on Cityscapes~\cite{cordts2016cityscapes}. We use their default 2e-5 learning rate, 39 layers, batch size of 2, and official XL 22K pretrained model for our base, CLIP, adverse only, and ablation models.

\subsection{ConvNeXt}
We use ConvNeXt's config for their XL model from their official Github repository. We maintain their default 4e-5 learning rate, 36 layers, batch size of 2, and official XL 22k pretrained model for our base, adverse only, and CLIP versions.

\subsection{Swin Transformer}
We use Swin Transformer's config for their 12 window model from the official MMSegmentation implementation. We maintain their default 4e-5 learning rate, 24 layers, batch size of 2, and converted 12 window 22k pretrained model for our base and adverse only model variations. For our CLIP model variations, due to the shape of Swin Transformer's latent features, we have to use a batch size of 1 with 2 gradient accumulation steps. Moreover, we add the CLIP injection layers before each downsampling layer, which happens at the end of each stage excluding the last one. However, we also exclude the first stage from the CLIP injection, as Swin Transformer's architecture has a very large encoding at the end of the first stage. Therefore, adding a CLIP injection layer at that stage can greatly increase the memory usage above the capacity of most GPUs, which is why we exlcude it from that layer. Thus, we have 2 CLIP injection layers for our augmented Swin Transformer model. 

\begin{table}
  \centering
  \small
  \begin{tabular}{cc}
    \toprule
    Models & Links\\
    \midrule
    InternImage ~\cite{wang2023internimage} & \url{https://github.com/OpenGVLab/InternImage}\\
    ConvNeXt~\cite{liu2022convnet} & \url{https://github.com/facebookresearch/ConvNeXt}\\
    Swin~\cite{liu2021swin} & \url{https://github.com/open-mmlab/mmsegmentation/tree/main/configs/swin}\\
    \bottomrule
  \end{tabular}
  \caption{\textbf{Code links for the used foundational models.}}
  \label{tab:implementations}
\end{table}

\section{Failure Cases}
\label{sec:failure}
\begin{figure*}[t]
    \centering
    \includegraphics [width=\linewidth]{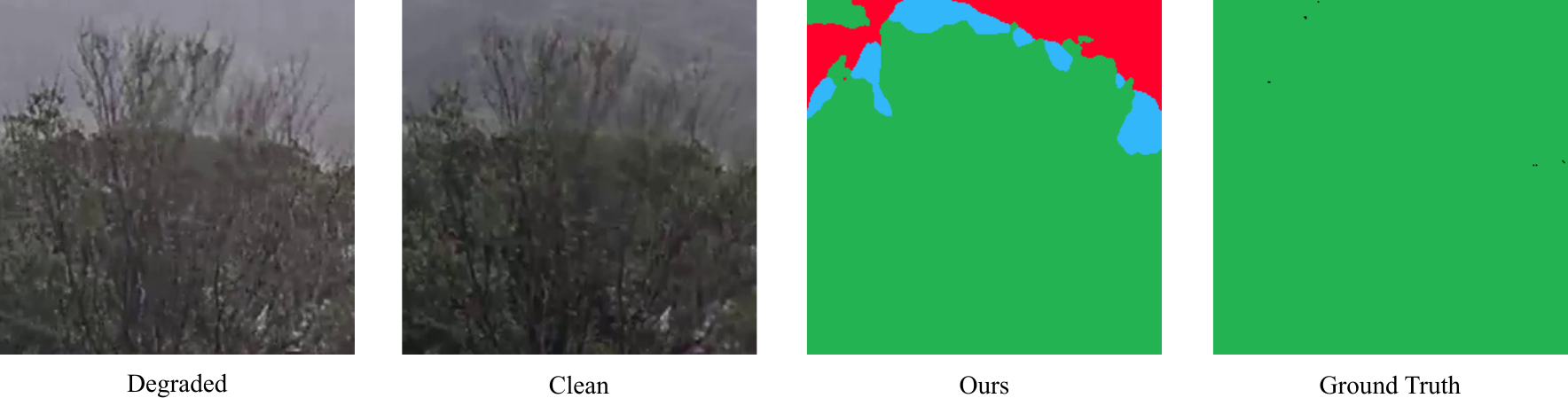}
    \caption{\textbf{Our paired training method does not perform well when there are occlusions that completely cover significant parts of a scene.} As shown in the figure, the background becomes masked from fog in the degraded image. As such, when training on an image like this, the consistency loss would confuse the model.}
    \label{fig:failures}
\end{figure*}

As seen in~\Cref{fig:failures}, our consistency losses and CLIP guidance would fail when the weather effects in an image are so significant that they fully obstruct parts of a scene. While this may be obvious for the shown scene, it is much harder to notice for weather effects that only show up in certain frames, such as large snow flakes that are close to the lens.

\end{document}